\documentclass[runningheads]{llncs}

\usepackage{eccv}

\usepackage{eccvabbrv}

\usepackage{graphicx}
\usepackage{booktabs}
\usepackage{makecell}
\usepackage{multirow}

\usepackage[accsupp]{axessibility}  

\usepackage{orcidlink}

\begin{document}

\title{KineBench: Benchmarking Embodied World Models via IDM-Free Kinematic Grounding}

\titlerunning{KineBench}

\author{Zeyu Liu\inst{1,2}\thanks{indicates equal contribution.} \and 
Zhangzhe Zhu\inst{1,3 \star} \and 
Yang Zhang\inst{1,4 \star}\thanks{Project Lead \\ Correspondence to: Chenjia Bai $<$baicj@chinatelecom.cn$>$} \and 
Chenyou Fan\inst{1,3} \\
Chenjia Bai\inst{1,5} \and 
Xuelong Li\inst{1}
}

\authorrunning{Zeyu Liu*, Zhangzhe Zhu*, Yang Zhang* (equal contribution)}

\institute{Institute of Artificial Intelligence (TeleAI), China Telecom, China
\and National University of Singapore, Singapore
\and Fudan University, China
\and Tsinghua University, China
\and Shenzhen Research Institute of Northwestern Polytechnical University, China}

\maketitle

\begin{abstract}
Evaluating the physical consistency of embodied world models (EWMs) is a critical open challenge.
While closed-loop evaluation via simulator rollouts offers a more faithful assessment of physical plausibility than open-loop alternatives, existing frameworks almost exclusively rely on Inverse Dynamics Models (IDMs) for action extraction.
Due to the intricate mapping from 2D pixel space to 3D kinematic space, the learned IDMs can be brittle to data outside their training distribution, resulting in unreliable action extraction from the generated videos with novel objects and scenarios.
This creates an unavoidable attribution ambiguity between world model inaccuracies and extractor errors.
To reduce this ambiguity, we present \textbf{KineBench}, an IDM-free closed-loop benchmark for EWMs, built upon an explicit kinematic grounding pipeline. Given a generated video, KineBench employs cascaded visual foundation models to directly extract 6D end-effector poses from individual frames, which are then executed in a physics simulator for closed-loop validation.
This explicit grounding directly tests the physical feasibility rather than visual plausibility, while remaining sensitive to general physical hallucinations such as gripper vanishing or spatial inconsistency.
Beyond execution-based task success, KineBench incorporates two classical 3D kinematic metrics—Spectral Arc Length (SPARC) and the Maruyama Manipulability Index—to characterize trajectory smoothness and kinematic feasibility from a robot-centric perspective. Across the evaluated models and tasks, these metrics exhibit task- and model-dependent associations with physical success rates, suggesting that they provide complementary diagnostic signals for assessing embodied generation quality.
Built on 20 diverse manipulation tasks in ManiSkill3, KineBench evaluates EWMs across four progressive suites: basic execution, task transfer, visual out-of-distribution generalization, and complexity-conditioned scaling.
Evaluation across frontier models reveals task-complexity-bounded nonlinear scaling in embodied video generation, providing empirical guidance for future data-scaling strategies.
The code and datasets are available on GitHub at
\url{https://github.com/minecraft-zzz/KineBench}
and on Hugging Face at
\url{https://huggingface.co/datasets/Zorkzak/KineBenchDatasets}.

\end{abstract}

\section{Introduction}\label{sec:intro}

\begin{figure}[htbp]
    \centering
    \includegraphics[width=1.0\textwidth]{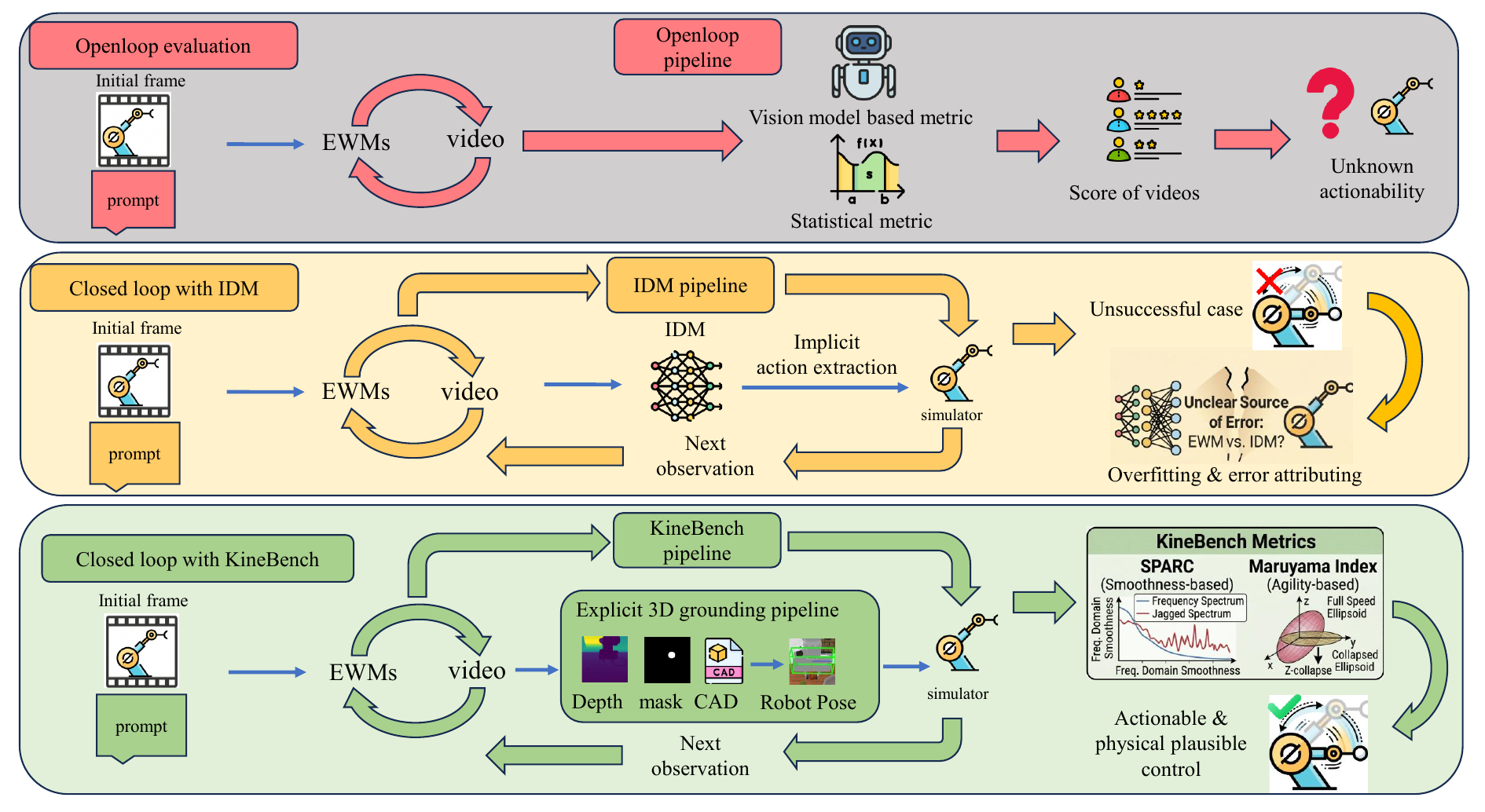}
     \caption{\textbf{Comparison of evaluation paradigms for Embodied World Models.}
KineBench provides an IDM-free, 3D kinematic grounding pipeline for rigorous physical validation, resolving issues of physical actionability and attribution ambiguity.}
    \vspace{-2.5em}
    \label{fig:three_evaluation_paradigms}
\end{figure}

Constructing Embodied World Models (EWMs) has emerged as a core objective in Embodied AI, aiming to endow agents with a predictive understanding of environmental dynamics~\cite{ha2018worldmodels, lecun2022path, liu2024sora, hafner2019dreamer, hafner2020dreamerv2, hafner2025dreamerv3, zhang2025marie, zhang2025dima}. Driven by recent advancements in scalable generative architectures, visual foundation models trained on internet-scale data have demonstrated remarkable spatiotemporal predictive capabilities~\cite{sora2024, wan2025wan, kong2024hunyuanvideo, wu2025hunyuanvideo15, seedance2025seedance15}. Consequently, an active line of research seeks to adopt these video generation models directly as predictive priors for physical intelligence and embodied control~\cite{chi2025wow, zhang2025world}.
These models serve two complementary roles in embodied control: as pre-training backbones for Vision-Language-Action models, where their richer spatiotemporal priors are hypothesized to yield stronger action policies~\cite{hu2024vpp, videovla2025, cen2025worldvla, ye2026dreamzero}; and as model-based planners that provide predictive rollouts to guide downstream task execution~\cite{chen2025large}.
However, this paradigm shift exposes a critical evaluation issue: \textit{how can we rigorously assess the actionable potential and physical plausibility of these video foundation models, rather than visual generation quality alone?}

Existing video generation benchmarks primarily evaluate pixel-level fidelity and spatiotemporal consistency~\cite{huang2024vbench, upadhyay2025worldbench}, which fail to capture the physical plausibility and kinematic feasibility required for embodied control.
Recent embodied benchmarks have gone further by leveraging vision models~\cite{ewmbench2025} and Large Vision-Language Models (VLMs)~\cite{deng2026rbench} to assess semantic correctness in an open-loop setting.
Beyond open-loop evaluation, a few frameworks attempt to ground generated sequences into physical simulation via simulators, providing a more direct assessment of executability~\cite{fan2026wow, zhang2025world}.
To close this loop in a model-agnostic manner, these frameworks almost exclusively have to rely on learned Inverse Dynamics Models (IDMs), which is either end-to-end neural mappings or explicit tracking-based approaches, to extract robot actions from generated RGB frames.
However, these closed-loop frameworks face a significant methodological bottleneck: how to extract robot actions from generated frames. Consequently, existing closed-loop benchmarks almost exclusively rely on Inverse Dynamics Models (IDMs)---either end-to-end neural mappings or explicit tracking-based approaches---to recover robot actions from RGB frames.

This IDM-based protocol, however, introduces a fundamental confounding factor. Due to the intricate mapping from 2D pixel space to 3D kinematic space, IDMs overfit the specific action distributions in their training data and frequently fail when faced with novel trajectories produced by generative models (see Sec.~\ref{sec:pipeline_analysis}).
This creates an unavoidable attribution ambiguity: when closed-loop execution fails, it is impossible to determine whether the failure stems from the world model generating physically implausible visuals, or simply from the IDM's poor generalization to unseen trajectories.

To reduce this attribution ambiguity, we introduce \textbf{KineBench}, an IDM-free closed-loop evaluation framework for EWMs based on explicit kinematic grounding. Rather than inferring actions through an implicit inverse dynamics model, KineBench recovers end-effector motion through a modular geometric pipeline. Figure~\ref{fig:three_evaluation_paradigms} compares three evaluation paradigms.
Given a language instruction and an initial observation, the world model generates a predictive video sequence. KineBench then applies a cascade of visual perception models to estimate the 6D end-effector pose from each generated frame, without requiring any modification to the evaluated world model. The recovered pose sequence is subsequently executed in a physics simulator for closed-loop evaluation.
The CAD-constrained pose-tracking procedure also provides a geometric regularization effect: rigid-body constraints can suppress small frame-level jitters and local visual artifacts, while larger inconsistencies, such as gripper disappearance or discontinuous spatial motion, tend to produce unstable or infeasible pose estimates. In this sense, the procedure exhibits a low-pass filtering effect analogous to that of an admittance controller~\cite{landi2017admittance}, although its reliability remains dependent on the underlying segmentation, depth estimation, and pose-tracking modules.

Beyond execution-based assessment, KineBench provides a \textbf{robot-centric} evaluation perspective grounded in 3D kinematics. Rather than assessing generated videos solely through visual appearance, we characterize the physical properties of the generated motions in 3D space. Specifically, KineBench incorporates \emph{Spectral Arc Length (SPARC)} to measure trajectory smoothness in the frequency domain and the \emph{Maruyama Manipulability Index} to characterize kinematic dexterity and feasibility under the robot model.

Across the evaluated models and tasks, these metrics exhibit task- and model-dependent associations with simulator execution success, suggesting that trajectory smoothness and kinematic feasibility provide complementary signals for assessing embodied generation quality. To the best of our knowledge, KineBench is the first framework to incorporate these classical robotics metrics into the evaluation of generative video models.

To systematically probe the capability boundaries of EWMs, we design 20 diverse manipulation tasks in ManiSkill3, organized into four evaluation suites of increasing diagnostic depth: basic execution capability, task-level transfer, visual out-of-distribution generalization, and complexity-conditioned scaling.

Across the evaluated frontier models, we observe complexity-conditioned nonlinear scaling in embodied video generation. In contrast to the broadly reported scaling trends of visual and language foundation models~\cite{kaplan2020scaling}, the marginal benefits of increasing data and compute diminish as task difficulty rises, providing empirical evidence to inform future scaling strategies for embodied world models.

In summary, our main contributions are as follows:
\begin{itemize}
    \item \textbf{IDM-free closed-loop benchmark.} We present \textbf{KineBench}, a closed-loop benchmark for EWMs that connects generated videos to physics simulation through an explicit 6D end-effector pose grounding pipeline. By replacing learned IDM-based action extraction with modular geometric estimation, KineBench reduces the attribution ambiguity between world-model errors and action-extraction errors.
    \item \textbf{Robot-centric kinematic evaluation paradigm.} We incorporate SPARC and the Maruyama Manipulability Index into generative video assessment. Across the evaluated models and tasks, these metrics exhibit task- and model-dependent associations with execution success, providing complementary diagnostic signals for embodied generation quality.
    \item \textbf{Structured four-suite benchmark and scaling analysis.} We design 20 diverse manipulation tasks in ManiSkill3, organized into four evaluation suites, covering basic execution, task transfer, visual OOD generalization, and complexity-conditioned scaling. Experiments across frontier models show that the marginal benefits of increasing data and compute diminish as task difficulty rises, providing empirical evidence to inform future scaling strategies for EWMs.
\end{itemize}

\vspace{-0.5em}
\section{Related Work}
\subsection{Video Models as Embodied World Models}
Large-scale video foundation models have demonstrated remarkable spatiotemporal predictive capabilities, learning rich physical dynamics from internet-scale data~\cite{sora2024, wan2025wan, kong2024hunyuanvideo}. This has catalyzed a novel research paradigm: adopting video generation models as world models for embodied agents~\cite{du2023unipi, yang2023unisim, bruce2024genie}. Under this paradigm, these models serve as predictive priors for robots, autoregressively imagining future physical interaction sequences conditioned on current observations and language instructions. Several works have further explored using such models as pre-training backbones for Vision-Language-Action models~\cite{hu2024vpp, videovla2025, cen2025worldvla,ye2026dreamzero}, hypothesizing that richer physical priors yield stronger downstream action policies. Despite these advances, rigorously evaluating their generation quality and physical plausibility remains an open challenge in the field.

\subsection{Benchmarks for Embodied World Models}
\textbf{Open-loop visual benchmarks.} Early video benchmarks primarily assess pixel-level fidelity and aesthetic quality~\cite{unterthiner2019fvd, huang2024vbench, qiao2025vadb}. Recent embodied-specific benchmarks have gone further: EWMBench~\cite{ewmbench2025} proposes a dedicated framework evaluating visual scene consistency, motion correctness, and semantic alignment for robotic manipulation via VLM-based assessment.
RBench~\cite{deng2026rbench} introduces a large-scale evaluation covering five task domains and four robot embodiments, benchmarking 25 models with metrics including structural consistency and action completeness.
These benchmarks represent significant progress but remain confined to open-loop, pixel-space evaluation, unable to verify physical feasibility.

\textbf{Physics-grounded open-loop benchmarks.} To measure physical plausibility without physical simulation, existing approaches formulate physics-grounded prompt suites and employ multimodal evaluators to decouple semantic adherence from physical commonsense~\cite{meng2025phygenbench, bansal2025videophy}.
Advanced paradigms bypass external judges by leveraging surrogate likelihoods from diffusion processes or extracting verifiable optical flow proxies to enforce Newtonian kinematics via reward constraints~\cite{yuan2025likephys, le2025newtonrewards}. However, these metrics still measure visual plausibility rather than actual physical feasibility.

\textbf{Closed-loop execution benchmarks.} A few benchmarks ground generated sequences into physical simulation for closed-loop evaluation~\cite{qin2024worldsimbench, zhang2025world, fan2026wow}.
World-in-World~\cite{zhang2025world} introduces a unified closed-loop interface and demonstrates that visual quality does not reliably correlate with task success---underscoring the necessity of execution-based evaluation.
WoW-World-Eval~\cite{fan2026wow} further proposes an IDM-based Turing Test to assess whether generated videos can fool a dynamics model trained on real trajectories.
However, all existing closed-loop benchmarks rely on IDMs for action extraction, inheriting the generalization limitations we identify in Sec.~\ref{sec:pipeline_analysis}. KineBench is the first to entirely eliminate this dependency via explicit kinematic grounding.

\vspace{-0.5em}
\subsection{Evaluation Metrics: From Pixels to 3D Kinematics}
Early video quality metrics measure frame-level pixel distribution distances or structural similarities~\cite{unterthiner2019fvd, wang2004ssim}. Subsequently, metrics based on 2D point tracking or optical flow were introduced to capture temporal consistency~\cite{huang2024vbench}. For real-world robotic manipulation, however, movement smoothness and kinematic dexterity are the core indicators of robust control.
Spectral Arc Length (SPARC)~\cite{balasubramanian2011sparc} has been widely validated in clinical and robotic settings as a more robust smoothness metric than traditional jerk-based measures, as it is insensitive to high-frequency noise.
The Yoshikawa and Maruyama Manipulability Index~\cite{yoshikawa1985manipulability} evaluates a robot's ability to avoid kinematic singularities and exert multi-directional forces across configurations.
To the best of our knowledge, KineBench is the first work to introduce these classical 3D kinematic metrics into the evaluation of generative video models.

\vspace{-0.5em}
\section{Methodology}\label{sec:method}
To evaluate the actionable capabilities and physical consistency of EWMs, we introduce \textbf{KineBench}, an IDM-free closed-loop evaluation framework comprising three components: (1) an explicit kinematic grounding pipeline that extracts 6D end-effector poses directly from generated frames; (2) a robot-centric kinematic evaluation system grounded in 3D metrics; and (3) a four-suite benchmark of 20 diverse manipulation tasks in ManiSkill3.

\begin{figure}[htbp]
    \centering
    \includegraphics[width=1.0\textwidth]{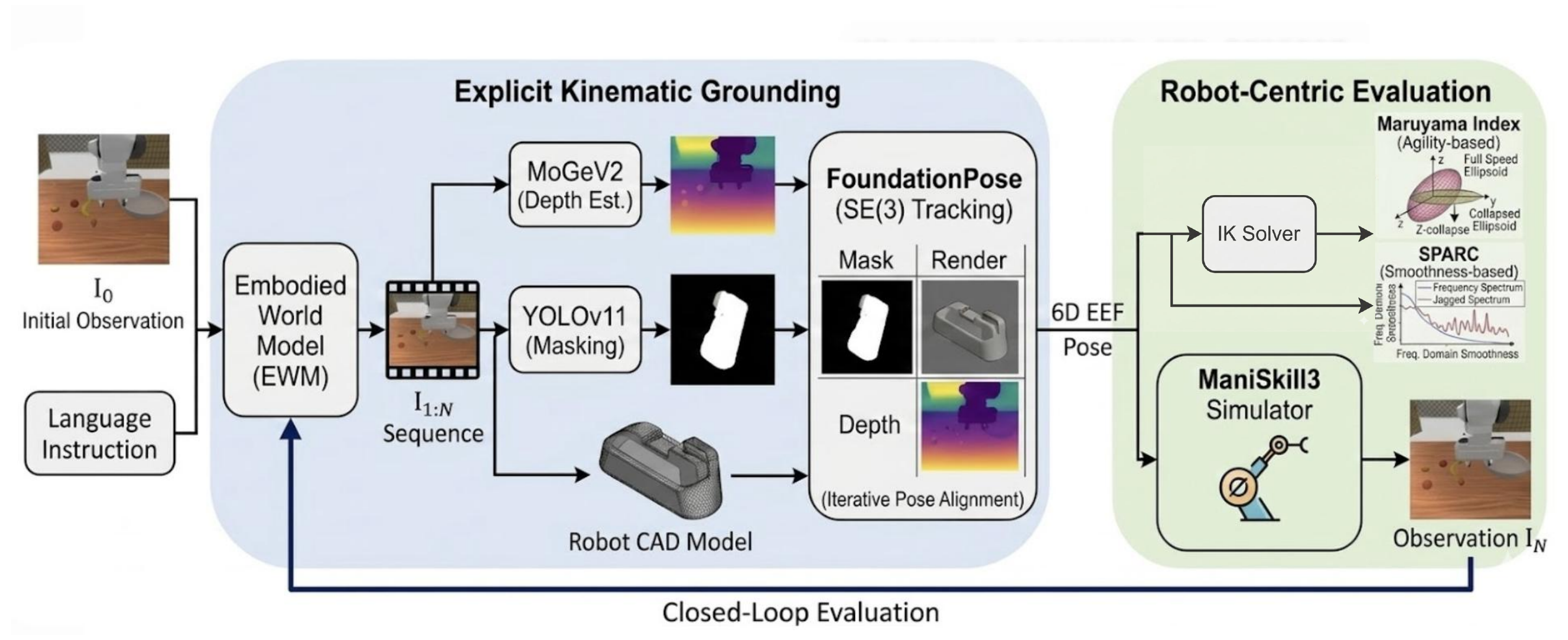}
    \caption{KineBench evaluation pipeline}
    \label{fig:kineBench_evaluation_pipeline}
    \vspace{-2.5em}
\end{figure}
\subsection{Explicit Kinematic Grounding Pipeline}
The inherent generalization limitations of IDMs introduce significant confounding factors into closed-loop evaluation, as discussed in Sec.~\ref{sec:intro}. To reduce this bottleneck, KineBench constructs a pipeline integrating 2D instance segmentation, monocular depth estimation (MoGeV2~\cite{wang2025moge2}), and 6D pose tracking (FoundationPose~\cite{wen2024foundationpose}). The overall KineBench pipeline is illustrated in Figure~\ref{fig:kineBench_evaluation_pipeline}

\begin{itemize}
\item \textbf{Masking and Depth Recovery.} Prior to 6D pose extraction, accurate 2D localization and 3D spatial grounding are required. We employ a fine-tuned YOLO model~\cite{ultralytics2025yolo} to segment the end-effector mask. Since the exact CAD models of the gripper are readily available from manufacturers, no manual annotation is required. We then apply a fine-tuned MoGeV2 within the simulation domain to produce high-precision absolute metric depth maps.

\item \textbf{CAD-based Pose Matching and Geometric Filtering.}
Given the segmentation mask, CAD model, and depth map, we apply FoundationPose to extract the 6D pose of the gripper. Its core mechanism follows a Render-and-Compare paradigm: the pose within $SE(3)$ space is iteratively refined by minimizing the photometric and geometric residuals between rendered CAD templates and the generated observation frames.
\end{itemize}

Crucially, while generated videos inevitably exhibit high-frequency temporal jitter or local non-rigid pixel deformations, FoundationPose enforces strict alignment with rigid-body constraints in 3D space. Consequently, this end-effector-centric approach naturally absorbs negligible non-physical pixel noise while remaining acutely sensitive to genuine physical hallucinations, e.g., gripper vanishing or extreme spatial inconsistencies that fundamentally violate rigid-body mechanics. The extracted 6D poses are then injected into ManiSkill3, which executes the actions, renders the next observation, and initiates the subsequent generation-and-rollout cycle.

\subsection{Robot-Centric Kinematic Evaluation}

Traditional video generation metrics, such as PSNR and FVD~\cite{unterthiner2019fvd}, operate exclusively within the 2D pixel domain, making them inadequate for assessing the physical plausibility of robotic motion. Because KineBench explicitly recovers 3D end-effector poses, we can shift the evaluation paradigm from visual appearance to kinematic fidelity. To this end, we incorporate two classical 3D kinematic metrics to measure trajectory smoothness and spatial dexterity, establishing a direct standard for evaluating an embodied world model's physical understanding.

\begin{itemize}
\item \textbf{Spectral Arc Length (SPARC).}
Traditional movement smoothness metrics rely on high-order temporal derivatives of position signals, which disproportionately amplify the minute high-frequency pose noise in video generation, causing the metrics invalid. SPARC resolves this issue by shifting the analytical domain to the frequency domain. It is defined as the arc length of the normalized Fourier magnitude spectrum:
$$\text{SPARC} \triangleq -\int_0^{\omega_c} \left[ \left(\frac{1}{\omega_c}\right)^2 + \left( \frac{d\hat{V}(\omega)}{d\omega} \right)^2 \right]^{\frac{1}{2}} d\omega$$
where $\hat{V}(\omega)$ is the normalized velocity magnitude spectrum, and $\omega_c$ is the adaptive cutoff frequency. Physically, smooth actions adhering to the dynamic principle of minimum energy concentrate their energy in low frequencies, resulting in a flat spectrum curve with a SPARC value approaching 0; conversely, if the model generates hallucinations such as stuttering or teleportation, high-frequency harmonics increase, the curve becomes convoluted, and the SPARC value significantly decreases. In this context, SPARC emerges as an exceptional standard for scrutinizing the temporal consistency of world models.
\item \textbf{Maruyama Manipulability Index.}
Smoothness alone is insufficient to guarantee manipulation success; robots must persistently avoid entering singular configurations. The manipulability index $w = \sqrt{\det(J(q)J^T(q))}$, based on the determinant of the Jacobian matrix $J(q)$, characterizes the volume of the end-effector's velocity ellipsoid. By inputting the video-generated 6D poses into an inverse kinematics (IK) solver, if the poses generated by the world model exceed the physical joint constraints of the robot, the manipulability index will sharply decline. Therefore, the integral of manipulability over the entire trajectory directly reflects the world model's implicit cognition of the robot's kinematic boundaries.
\end{itemize}

\subsection{Four-Suite Benchmark Design}
\begin{figure}[htbp]
    \centering
    \includegraphics[width=0.95\textwidth]{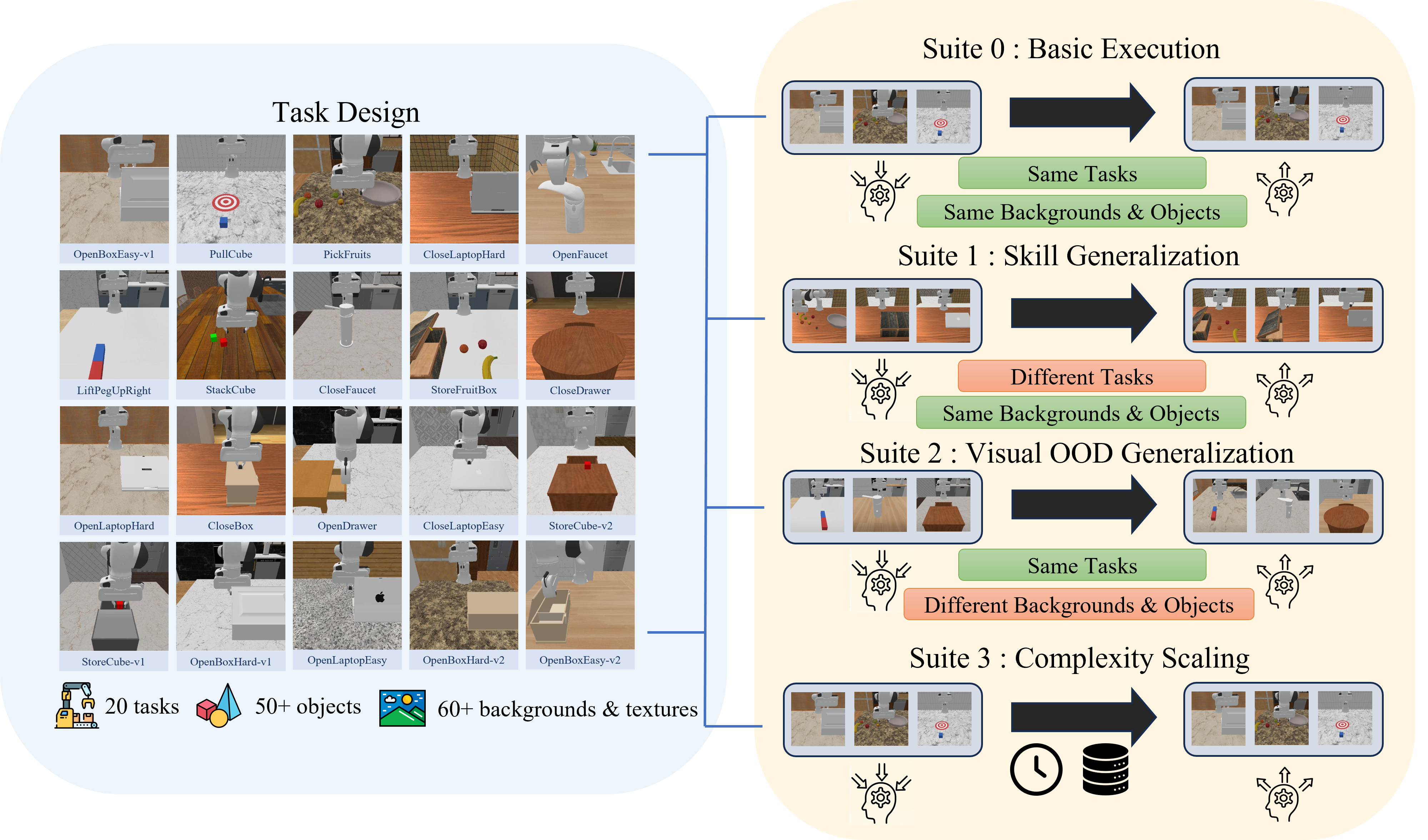}
    \caption{\textbf{Overview of the KineBench Task and Benchmark Design}}
    \label{fig:benchmark_suites}
\end{figure}
To systematically probe the capability boundaries of EWMs, we design 20 diverse manipulation tasks in ManiSkill3, spanning a wide spectrum of physical complexities from primitive rigid-body interactions (e.g., pick-and-place) to long-horizon articulated object manipulation\footnote{Part of the assets used in our tasks are derived from the HumanoidGen project \cite{jing2026humanoidgen}. We thank the HumanoidGen team for their valuable contributions.}. To isolate specific axes of generalization, these tasks are organized into four evaluation suites:

\paragraph{Suite 0: Basic Execution.} The training and evaluation sets are strictly identically distributed (I.I.D.), encompassing all tasks, object instances, and backgrounds. This suite establishes the fundamental kinematic execution baseline, quantifying a model's capacity to reliably reproduce demonstrated dynamics without domain shift.

\paragraph{Suite 1: Task Transfer.} To test whether models digest physical causality rather than merely memorizing trajectory distributions, this suite evaluates zero-shot transfer to conceptually related but disjoint tasks. For instance, an EWM trained on \texttt{CloseBox}, \texttt{OpenBoxEasy-v1}, and \texttt{CloseLaptopEasy} is evaluated on unseen spatial recombinations like \texttt{OpenBoxEasy-v2} (novel box orientations) or harder variants like \texttt{CloseLaptopHard}. This suite probes the model's capacity for instruction understanding, spatial extrapolation and physical causality reasoning.

\paragraph{Suite 2: Visual OOD Generalization.} This suite decouples visual representation robustness from physical reasoning. The training set is restricted to 50\% of the randomized asset library (object instances and background textures), while evaluation is conducted entirely on the disjoint remaining 50\%. This strictly out-of-distribution (OOD) visual setting tests whether kinematic generation catastrophically degrades under novel textures, geometries, and lighting conditions.

\paragraph{Suite 3: Complexity Scaling.}
Following the I.I.D. task distribution of Suite 0, we conduct controlled scaling experiments by systematically varying the training data volume and compute budget, with training duration used as a proxy for compute. This suite examines how intrinsic task complexity, such as degrees of freedom and contact dynamics, affects the performance gains obtained from increased data and compute, thereby revealing complexity-conditioned empirical scaling trends for EWMs.

\vspace{-0.5em}
\section{Experiments}

\subsection{Experimental Setup}

\subsubsection{Evaluation Environments and Task Taxonomy}
We systematically evaluate the models using the ManiSkill3 physics simulator. As detailed in Figure~\ref{fig:benchmark_suites}, we meticulously design 20 diverse manipulation tasks encompassing a wide spectrum of physical challenges: basic reaching motions, rigid body grasping (with and without obstacle interference), multi-axis articulated object manipulations (e.g., hinges and sliders oriented along the X, Y, and Z axes), and long-horizon compositional tasks. To rigorously assess generalization, each task incorporates highly randomized object instances and background textures, and is further stratified into different difficulty levels.

\subsubsection{Hardware Constraints and Model Pool}
Constrained by a realistic computing budget of four NVIDIA A100 GPUs, we benchmark a representative suite of state-of-the-art video foundation models, including Wan 2.1 (1.3B), Wan 2.2 (5B), CogVideoX (2B), as well as LoRA-finetuned variants (CogVideoX 5B-LoRA and Wan 2.2 5B-LoRA).

\subsubsection{Implementation Details of the KineBench Pipeline}
For explicit action extraction, we utilize a fine-tuned YOLOv11 for 2D mask extraction, a two-stage fine-tuned MoGeV2 which amplifies the loss on fine-grained geometric details on the second stage for high-fidelity metric depth estimation, and FoundationPose for zero-shot 6D pose tracking. This decoupled extraction-and-rollout pipeline ensures a highly efficient and stable closed-loop evaluation.
\begin{figure}[htbp]
\centering

\begin{subfigure}[b]{0.72\textwidth}
    \centering
    \includegraphics[width=\linewidth]{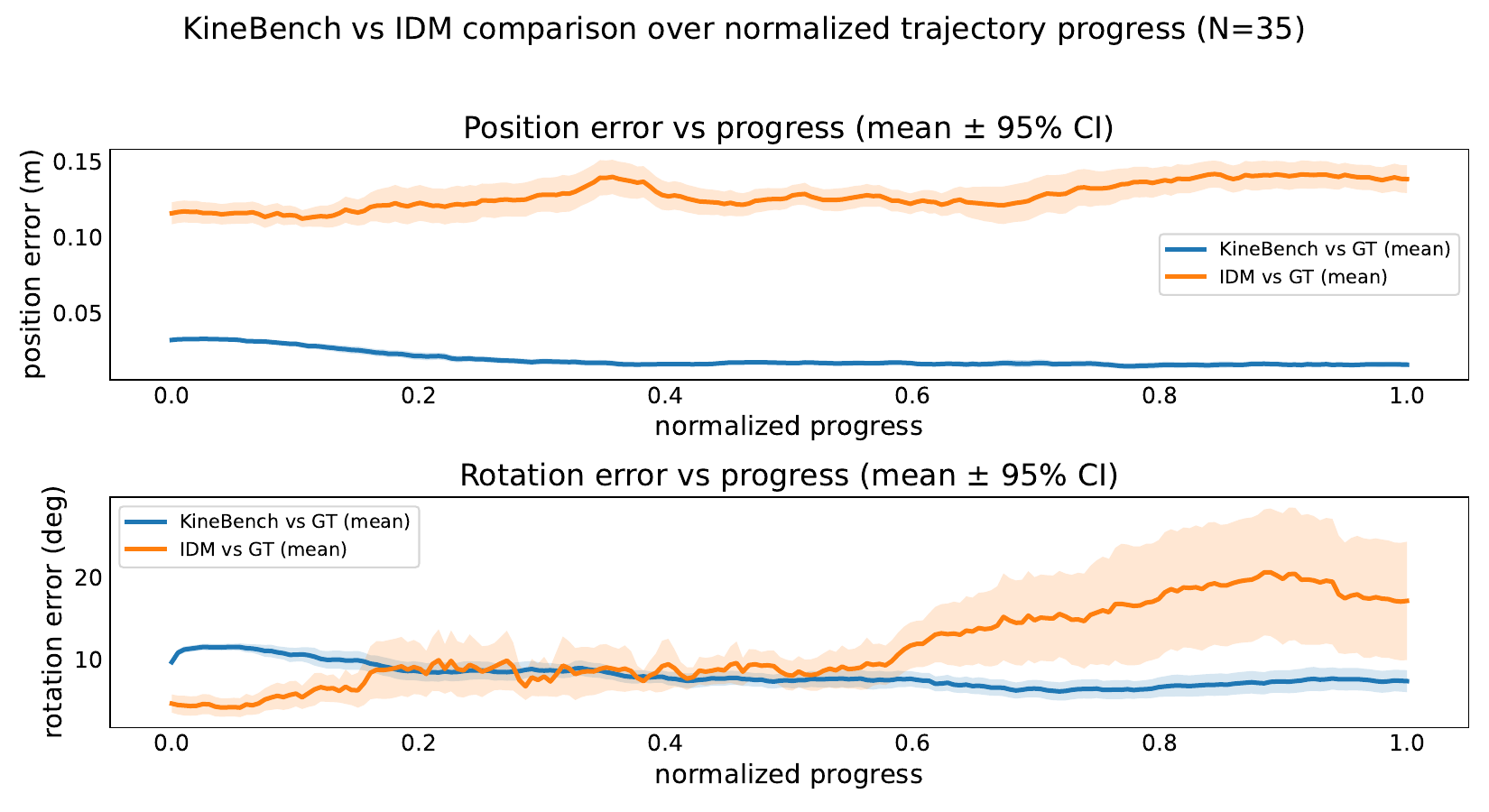}
    \caption{Trajectory error comparison}
    \label{fig:kinebench_idm_trajectory}
\end{subfigure}%
\begin{subfigure}[b]{0.23\textwidth}
    \centering
    \includegraphics[width=\linewidth]{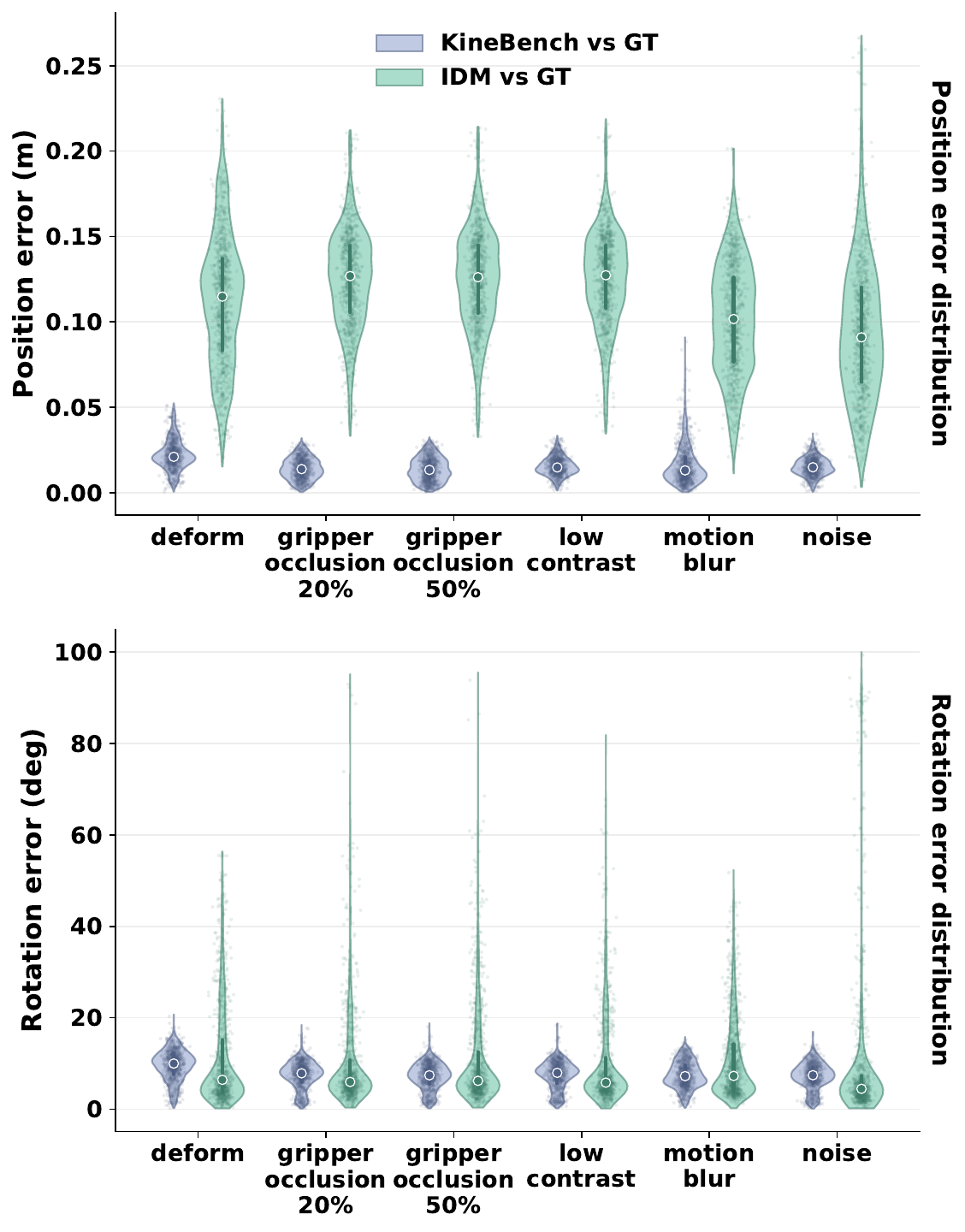}
    \caption{Distribution error comparison}
    \label{fig:kinebench_idm_perturbation}
\end{subfigure}

\caption{\textbf{Generalization and robustness comparison between KineBench and the IDM baseline.}
\textbf{(a)} Trajectory-wise translational and rotational errors on seven unseen test tasks.
The IDM is trained on all trajectories from the ten training tasks in Suite~1, with 100 trajectories per task.
For evaluation, we randomly sample five trajectories from each unseen task and compare the recovered end-effector poses with simulator ground-truth poses.
\textbf{(b)} Distributions of pose-estimation errors under controlled visual perturbations, including gripper deformation, increasing levels of occlusion, reduced contrast, motion blur, and image noise.
Lower values indicate more accurate pose recovery.}
\label{fig:KineBench_vs_IDM}
\end{figure}

\subsection{Generalization and Robustness Analysis of the KineBench Pipeline}
\label{sec:pipeline_analysis}

A central design choice of KineBench is to replace implicit IDM-based action inference with an explicit and modular geometric grounding pipeline. Although the pipeline still contains learned perception components, including segmentation and depth estimation, its intermediate outputs can be independently inspected and ablated. Our goal is therefore not to treat the extracted poses as ground truth, but to examine whether explicit geometric grounding provides a more stable and diagnosable interface than an IDM when evaluating out-of-distribution generated videos. We conduct controlled analyses covering task-level generalization, robustness to visual degradations, and sensitivity to depth estimation.

\subsubsection{Generalization and Robustness of Action Extraction}

We compare KineBench with Dino3DFlowIDM~\cite{fan2026wow} on simulator-rendered trajectories from tasks excluded from IDM training. Both methods receive the same visual observations, and their recovered end-effector poses are evaluated against simulator ground truth. We further introduce controlled perturbations that approximate common generated-video artifacts, including gripper deformation, varying levels of occlusion, reduced contrast, motion blur, and image noise. As shown in Figures~\ref{fig:kinebench_idm_trajectory} and~\ref{fig:kinebench_idm_perturbation}, across all conditions, the KineBench pipeline achieves substantially lower position error than the IDM baseline. Although rotation mean error is slightly higher in some cases, its distribution is much more concentrated, with far fewer long-tail failures. These results support the reliability of our pipeline and indicate that it provides a more robust evaluation interface than IDM.

\begin{figure}[htbp]
\centering

\begin{subfigure}[b]{0.70\textwidth}
    \centering
    \includegraphics[width=\linewidth]{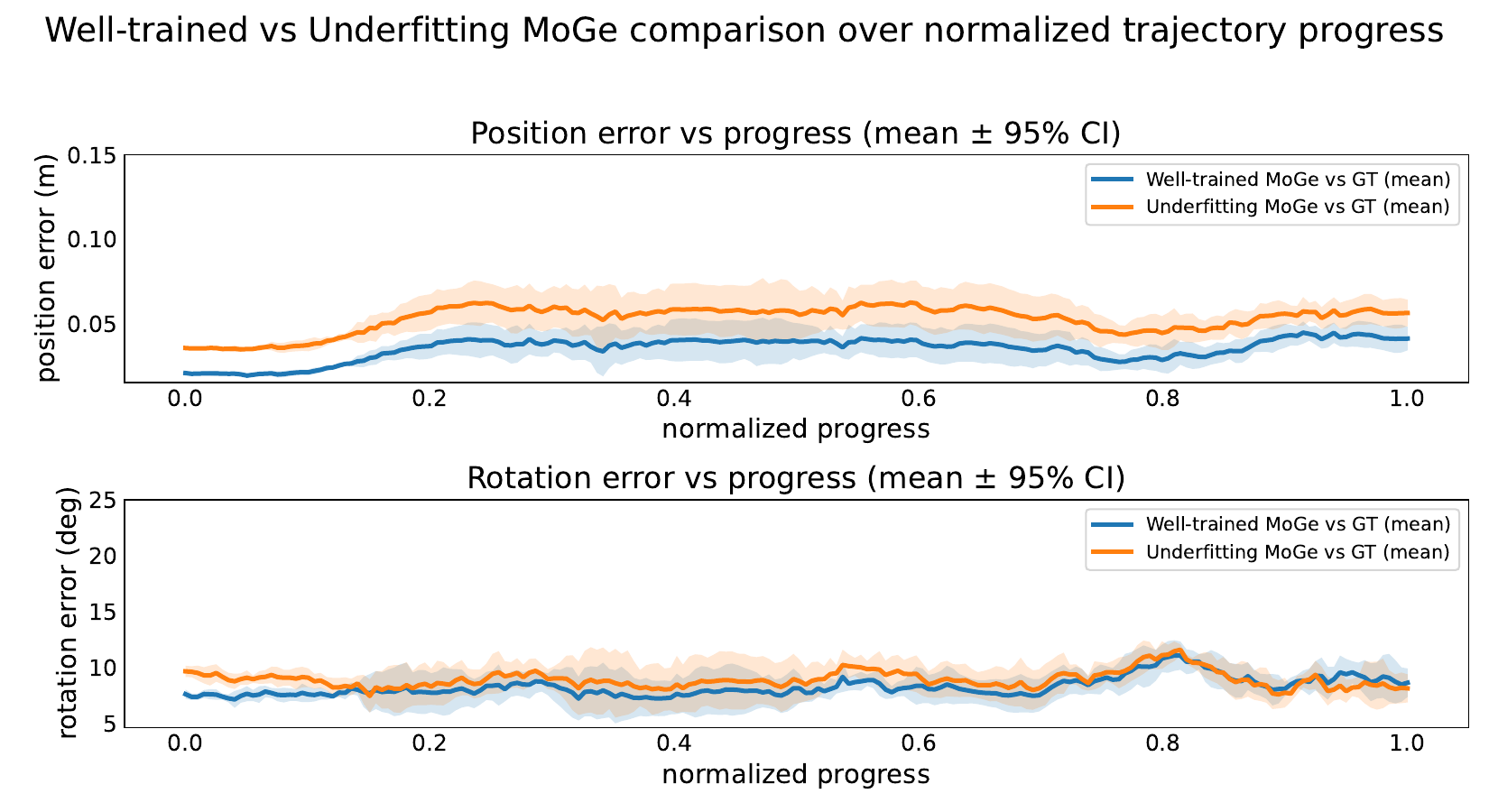}
    \caption{Trajectory error comparison}
\end{subfigure}%
\begin{subfigure}[b]{0.25\textwidth}
    \centering
    \includegraphics[width=\linewidth]{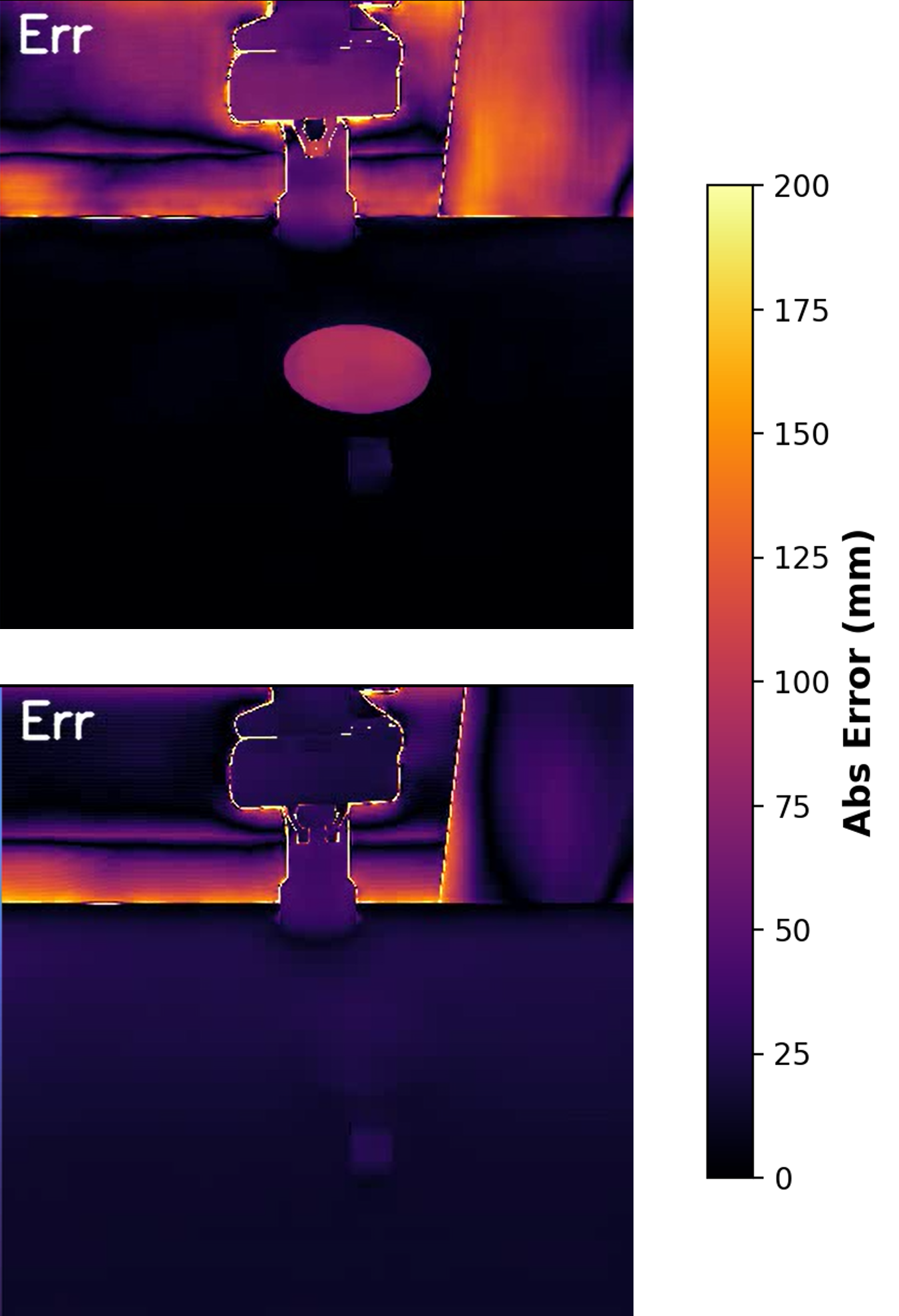}
    \caption{Depth error heatmap}
\end{subfigure}
\caption{\textbf{Comparison of pose and depth estimation errors under different MoGe depth quality levels.}
(a) Pose errors between trajectories reconstructed from MoGe-estimated depths and the ground-truth poses. 
(b) Heatmap visualization of depth errors between MoGe-estimated depths and the corresponding ground-truth depths. 
Despite the difference in depth estimation quality, the resulting pose errors remain very similar, demonstrating the robustness of the proposed pipeline.}
\label{fig:moge_comparison}
\vspace{-2.5em}
\end{figure}

\subsubsection{Ablation of Depth Estimation and Pose Tracking}

To isolate the effect of upstream depth estimation, we further compare three action-extraction settings: FoundationPose with simulator ground-truth depth, FoundationPose with MoGeV2-estimated depth, and Dino3DFlowIDM. Using ground-truth depth yields approximately centimeter-level translational errors, while replacing it with MoGeV2 increases the error to roughly 1.5--3\,cm. Figure~\ref{fig:moge_comparison} further examines sensitivity to depth-estimation quality. In comparison, the IDM exhibits errors on the order of ten centimeters on the evaluated unseen trajectories. The remaining rotational error of the MoGeV2-based pipeline is approximately ten degrees, indicating that orientation estimation remains an important limitation for precision manipulation.

This ablation suggests that most of the translational error introduced by the KineBench pipeline can be localized to the depth-estimation stage, while the CAD-constrained pose tracker remains comparatively stable. More importantly, the modular structure makes these error sources directly measurable, unlike an end-to-end IDM whose failures cannot be easily separated into perception and action-inference components.

Overall, the results indicate that KineBench provides a more robust and interpretable action-extraction interface under the evaluated distribution shifts and visual perturbations. However, its estimates remain dependent on segmentation, depth recovery, and sufficient visibility of the end-effector; accordingly, we view the pipeline as reducing IDM-specific attribution ambiguity rather than providing error-free pose recovery.

\subsection{Main Results: Physical Execution and Generalization}

\begin{table*}[t]
    \centering
    \caption{\textbf{Closed-loop success rates across evaluation suites.}
    All values are reported as percentages.
    ``Seen'' and ``Unseen'' correspond to the training and validation
    visual assets in Suite~2, respectively.
    A dash indicates that the model configuration was not evaluated in
    the corresponding suite.}
    \label{tab:suite_success_rates}

    \setlength{\tabcolsep}{2.4pt}
    \renewcommand{\arraystretch}{1.35}
    \setlength{\aboverulesep}{0.5ex}
    \setlength{\belowrulesep}{0.5ex}
    \scriptsize

    \resizebox{\textwidth}{!}{
    \begin{tabular}{lccccccccccccc}
        \toprule
        \multirow{3}{*}{\textbf{Suite}}
        & \multicolumn{7}{c}{\textbf{Main model comparison}}
        & \multicolumn{6}{c}{\textbf{Wan2.1-1.3B scaling configurations}} \\
        \cmidrule(lr){2-8}
        \cmidrule(lr){9-14}

        & \makecell{\textbf{Wan2.2}\\\textbf{5B}\\7.5k}
        & \makecell{\textbf{CogVideoX}\\\textbf{2B}\\7.5k}
        & \makecell{\textbf{Wan2.1}\\\textbf{1.3B}\\7.5k}
        & \makecell{\textbf{Wan2.2}\\\textbf{5B LoRA}\\7.5k}
        & \makecell{\textbf{CogVideoX}\\\textbf{5B LoRA}\\7.5k}
        & \makecell{\textbf{Wan2.6}\\API}
        & \makecell{\textbf{Hailuo-V2}\\API}
        & \makecell{\textbf{1.5k}\\steps}
        & \makecell{\textbf{4.5k}\\steps}
        & \makecell{\textbf{15k}\\steps}
        & \makecell{\textbf{Scale 10}\\7.5k}
        & \makecell{\textbf{Scale 25}\\7.5k}
        & \makecell{\textbf{Scale 50}\\7.5k} \\
        \midrule

        \textbf{Suite 0}
        & \textbf{56.32}
        & 46.17
        & 43.96
        & 19.00
        & 18.33
        & 14.08
        & 6.56
        & --
        & --
        & --
        & --
        & --
        & -- \\
        \addlinespace[2pt]

        \textbf{Suite 1}
        & 11.90
        & \textbf{57.78}
        & 20.00
        & 16.67
        & 3.81
        & --
        & --
        & --
        & --
        & --
        & --
        & --
        & -- \\
        \addlinespace[2pt]

        \textbf{Suite 2 (Seen)}
        & \textbf{58.00}
        & 56.11
        & 57.19
        & 25.33
        & --
        & --
        & --
        & --
        & --
        & --
        & --
        & --
        & -- \\
        \addlinespace[2pt]

        \textbf{Suite 2 (Unseen)}
        & \textbf{55.50}
        & 51.67
        & 52.83
        & 26.33
        & --
        & --
        & --
        & --
        & --
        & --
        & --
        & --
        & -- \\
        \addlinespace[2pt]

        \textbf{Suite 3}
        & --
        & --
        & 43.96
        & --
        & --
        & --
        & --
        & 44.83
        & 47.88
        & \textbf{73.33}
        & 53.67
        & 52.00
        & 51.17 \\

        \bottomrule
    \end{tabular}
    }

\end{table*}

We rigorously evaluate the models across three dimensions: base capacity (S0), cross-task transfer (S1), and visual OOD robustness (S2). Table~\ref{tab:suite_success_rates} presents a performance comparison across evaluated models.

\subsubsection{Base Physical Execution (Suite 0)}. The results show that larger models excel on complex tasks demanding fine-grained manipulation or long-horizon sequencing (e.g., OpenBoxHard-v1, StoreCube-v2). However, severe performance drops on contact-rich tasks (e.g., StackCube, PickFruits) reveal that while video models capture kinematic priors, modeling complex friction and collision dynamics remains an open challenge. Under the matched Wan2.2-5B setting, full-parameter fine-tuning achieves higher performance than LoRA in S0, indicating that updating the full model may be beneficial for acquiring task-specific execution dynamics. However, this advantage does not hold consistently across the other evaluation suites.

\subsubsection{Task and Visual Generalization (Suites 1 \& 2)}

In Suite~2, models remain relatively robust on simple tasks such as CloseBox, but degrade substantially on tasks requiring precise affordance localization under visual shifts. For example, Wan~2.2 drops from 60.0\% to 30.0\% on OpenBoxHard when evaluated on unseen assets. This suggests that current models may rely on appearance-specific features rather than learning transferable 3D geometric affordances.

Suite~1 further shows that cross-task transfer remains challenging. Notably, the LoRA-tuned model outperforms its fully fine-tuned counterpart, suggesting a trade-off between task-specific execution and transferability. One possible explanation is that LoRA preserves more pretrained visual and motion priors, whereas full-parameter fine-tuning specializes more strongly to the training tasks. We treat this explanation as a hypothesis rather than a causal conclusion.

To complement visual evaluation, we analyze the extracted 6D trajectories using SPARC and the Maruyama Manipulability Index. Their task- and model-dependent associations with execution success indicate that they provide complementary diagnostic signals for trajectory smoothness and kinematic feasibility.

\begin{figure}[htbp]
    \centering
    \includegraphics[width=0.95\textwidth]{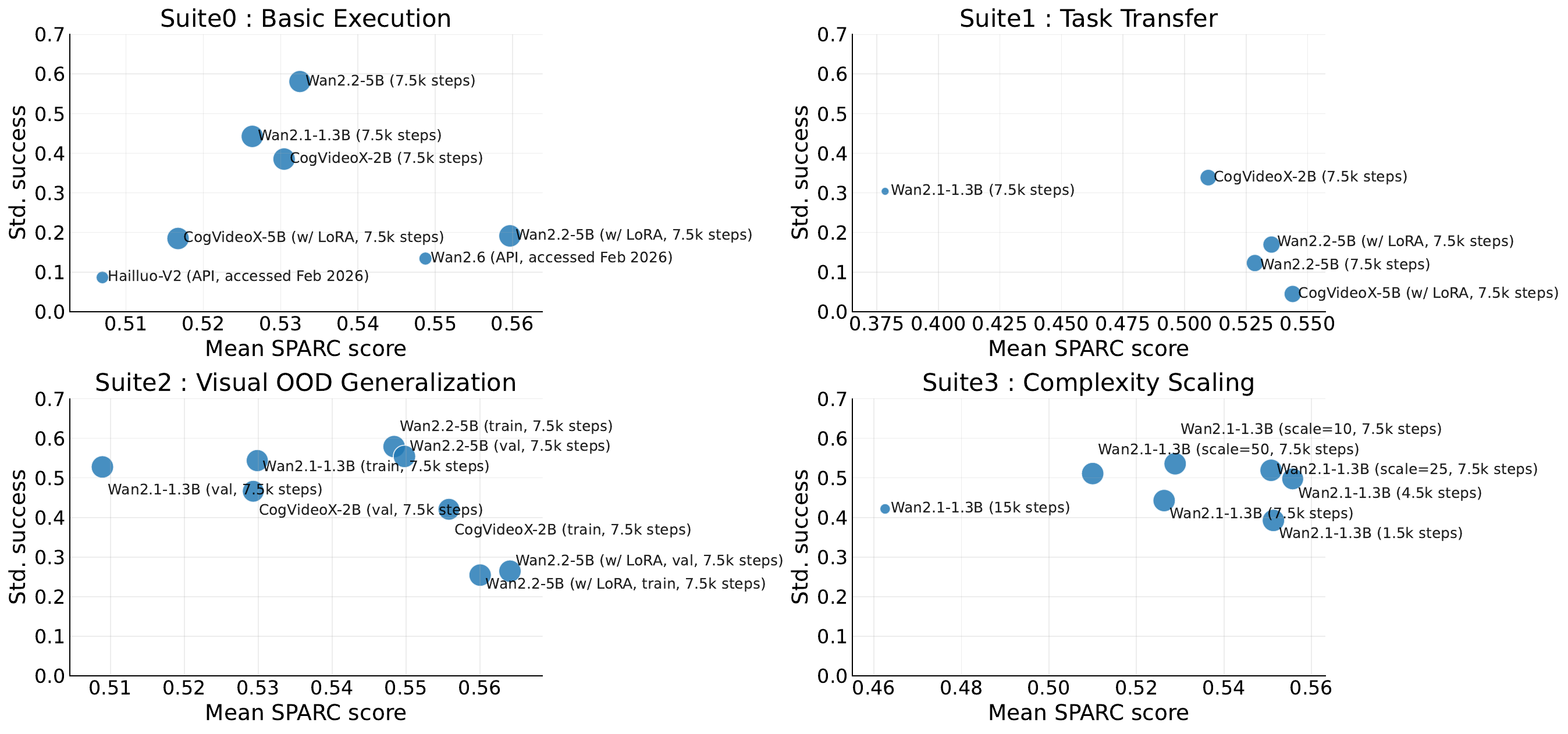}
    \caption{\textbf{Relationship between SPARC score and standardized success rate across models.} Each point represents one model within a benchmark suite. The x-axis shows the task-balanced mean SPARC score, and the y-axis shows the standardized success rate.}
    \label{fig:sparc}
    \vspace{-2.5em}
\end{figure}

\subsubsection{SPARC: Evaluating Model Motion Fluency}
SPARC’s predictive power is highly conditioned on a model's generation capabilities. Figure~\ref{fig:sparc} shows that for zero-shot and under-optimized models, this metric strongly correlates with success rates, effectively filtering catastrophic physical hallucinations caused by poor motion fluency. Conversely, fully fine-tuned models generate kinematically smooth trajectories that saturate SPARC scores; their failures arise primarily from semantic misalignments rather than local jitter. Therefore, SPARC serves as a rigorous, specialized metric for evaluating fine-grained temporal fluency in physical actions.

\subsubsection{Maruyama Manipulability: Evaluating Embodiment Awareness}
\begin{figure}[htbp]
    \centering
    \includegraphics[width=0.95\textwidth]{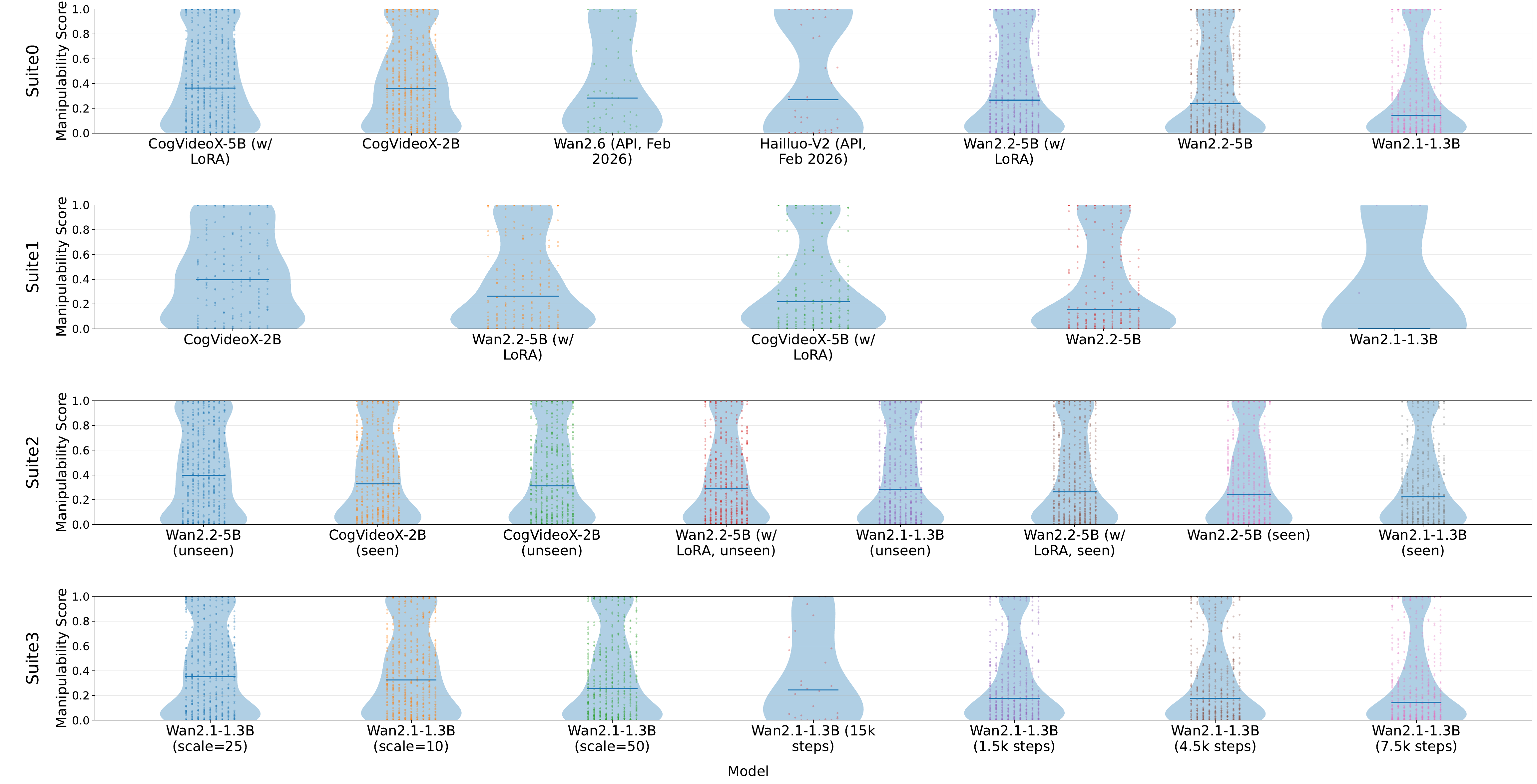}
    \caption{\textbf{Manipulation score across different suites.}
    The figure presents violin plots of the Manip score across different suites, illustrating the distribution of manipulation performance across models. 
    Lower values indicate better manipulation quality, corresponding to more accurate and stable execution of the generated trajectories. 
    The Manip score is computed from trajectory costs obtained via PyRoki and normalized using a robust min--max scaling based on the 10th--90th percentiles.}
    \label{fig:manip}
    \vspace{-2.5em}
\end{figure}

While SPARC captures temporal dynamics, the Maruyama Manipulability Index evaluates spatial constraints and embodiment awareness. Fully fine-tuned models effectively internalize the 7-DoF robot's structural boundaries, consistently maintaining low manipulability costs (Figure~\ref{fig:manip}). Conversely, zero-shot closed-source and capacity-limited LoRA models lack this familiarity. Although generating visually plausible motions, their inherent stochasticity frequently pushes the robot toward singularities or kinematically unreachable points, leading to execution failures. Consequently, this index serves as a crucial metric for embodiment familiarity, exposing the risks of unconstrained generative randomness in world models.

Together, these two metrics establish a comprehensive 3D evaluation paradigm: SPARC scrutinizes the dynamic motion details, while Manipulability bounds the spatial positioning and embodiment constraints, offering profound insights into why generated videos succeed or fail in the physical world.

\subsection{Complexity-Conditioned Scaling Behaviors}
In S3 of Table~\ref{tab:suite_success_rates}, we conduct a controlled scaling analysis using the Wan~2.1 1.3B architecture by varying the number of expert trajectories from 10 to 100 and the number of optimization steps from 1,500 to 15,000. Rather than exhibiting uniformly monotonic improvements, the results show task-dependent and nonlinear scaling trends.

For the relatively simple OpenBoxEasy task, increasing the number of homogeneous training trajectories from 10 to 100 reduces the success rate from 100\% to 83.33\%, while additional optimization steps also provide limited or negative gains. This suggests that increasing data and training duration does not necessarily improve performance when the task distribution is narrow and the model has already learned the dominant motion patterns. One possible explanation is that prolonged training on highly similar trajectories leads to overspecialization, although our experiments do not directly verify this mechanism.

In contrast, for the more complex StoreCube-v2 task, increasing the optimization budget improves the success rate from 13.34\% to 69.52\%, indicating that tasks with more diverse motion and interaction requirements may benefit from longer training. Overall, these results suggest that the marginal benefits of increasing data and compute depend on task complexity and trajectory diversity. Broader experiments across architectures, data regimes, and collection strategies are needed to determine how generally these trends hold.

\vspace{-0.5em}
\section{Discussion and Limitations}
While our benchmark evaluates several complementary aspects of embodied world models, the expert trajectories utilized in our benchmark are generated via motion planning. Future iterations will integrate human teleoperation data to enrich the action distribution. Although our 3D kinematic metrics effectively capture low-level physical coherence, future work will explore integrating them with existing 2D pixel-based evaluations, enabling a more holistic diagnosis of embodied world models.

\vspace{-0.5em}
\section{Conclusion}
In conclusion, we present KineBench, an IDM-free closed-loop benchmark for evaluating Embodied World Models through explicit 3D kinematic grounding. By incorporating classical robotic kinematic metrics, KineBench provides an executor-centric perspective on how selected properties of generated motions relate to downstream physical execution. Across four evaluation suites, our experiments reveal task-complexity-dependent nonlinear scaling trends among the evaluated video models. Overall, KineBench offers a transparent evaluation framework and empirical observations that may inform future research on physically grounded and actionable world models.


\section*{Acknowledgements}

This work is supported by the National Key Research and Development Program of China (Grant No.2024YFE0210900), the National Natural Science Foundation of China (Grant No.62306242),  the Young Elite Scientists Sponsorship Program by CAST (Grant No. 2024QNRC001), and the Yangfan Project of the Shanghai (Grant No.23YF11462200).

%
%
\bibliographystyle{splncs04}
\bibliography{main}
\end{document}